\documentclass{article}

\usepackage{latexsym}
\usepackage{amsthm}
\usepackage{enumitem}
\usepackage{color}

\usepackage[american]{babel}
\usepackage{arxiv}

\usepackage[utf8]{inputenc} 
\usepackage[T1]{fontenc}    
\usepackage{hyperref}       
\usepackage{url}            
\usepackage{booktabs}       
\usepackage{amsfonts}       
\usepackage{nicefrac}       
\usepackage{microtype}      
\usepackage{lipsum}		
\usepackage{graphicx}
\usepackage{natbib}
\usepackage{doi}
\usepackage{natbib} 

\usepackage{mathtools} 
\usepackage{tikz} 
\usepackage{algorithm}
\usepackage{algpseudocode}
\usepackage{amsmath,amssymb,amsfonts}
\usepackage{algcompatible}
\usepackage{multicol}
\usepackage{array,multirow,graphicx}
\usepackage{dblfloatfix}
\usepackage{caption}
\usepackage{subcaption}

\title{The Probabilistic Tsetlin Machine: A Novel Approach to Uncertainty Quantification\thanks{K. Darshana Abeyrathna, 2024. This is the author's version of the work. It is posted here for your personal use. Not for redistribution. The definitive version was published in \textit{2024 The 8th International Conference on Advances in Artificial Intelligence Proceedings}, \url{https://doi.org/10.1145/3704137.3704143}.}}

\author{ {K. Darshana Abeyrathna, Sara El Mekkaoui, Andreas Hafver and Christian Agrell}
\\
	Group Research and Development, DNV (Det Norske Veritas), Høvik, Norway
}

\renewcommand{\shorttitle}{Probabilistic Tsetlin Machine}

\hypersetup{
pdftitle={A template for the arxiv style},
pdfsubject={q-bio.NC, q-bio.QM},
pdfauthor={David S.~Hippocampus, Elias D.~Striatum},
pdfkeywords={Probabilistic Tsetlin Machine, Tsetlin Machine, Uncertainty Quantification},
}

\begin{document}
\maketitle 

\begin{abstract}
Tsetlin Machines (TMs) have emerged as a compelling alternative to conventional deep learning methods, offering notable advantages such as smaller memory footprint, faster inference, fault-tolerant properties, and interpretability. Although various adaptations of TMs have expanded their applicability across diverse domains, a fundamental gap remains in understanding how TMs quantify uncertainty in their predictions. In response, this paper introduces the Probabilistic Tsetlin Machine (PTM) framework, aimed at providing a robust, reliable, and interpretable approach for uncertainty quantification. Unlike the original TM, the PTM learns the probability of staying on each state of each Tsetlin Automaton (TA) across all clauses. These probabilities are updated using the feedback tables that are part of the TM framework: Type I and Type II feedback. During inference, TAs decide their actions by sampling states based on learned probability distributions, akin to Bayesian neural networks (BNNs) when generating weight values. In our experimental analysis, we first illustrate the spread of the probabilities across TA states for the noisy-XOR dataset. Then we evaluate the PTM alongside benchmark models using both simulated and real-world datasets. The experiments on the simulated dataset reveal the PTM's effectiveness in uncertainty quantification, particularly in delineating decision boundaries and identifying regions of high uncertainty. Moreover, when applied to multiclass classification tasks using the Iris dataset, the PTM demonstrates competitive performance in terms of predictive entropy and expected calibration error, showcasing its potential as a reliable tool for uncertainty estimation. Our findings underscore the importance of selecting appropriate models for accurate uncertainty quantification in predictive tasks, with the PTM offering a particularly interpretable and effective solution. 
\end{abstract}


\section{Introduction}\label{sec:intro}

Tsetlin Machines (TMs) are a promising alternative to current-day deep learning \cite{Ole1}. Initially, TMs were able to perform classification tasks with binary features, competitively against the state-of-the-art machine-learning approaches. Subsequently, different adaptations of TMs have been proposed to tackle a wide array of challenges across diverse fields. These adaptations cater to scenarios involving continuous inputs and outputs \cite{darshana2020regression}, deterministic learning processes \cite{abeyrathna2020novel}, natural language processing (NLP) applications \cite{saha2021using, bhattarai2022convtexttm}, image analysis tasks \cite{granmo2019convolutional}, as well as tasks related to compact pattern representation \cite{abeyrathna2021extending}.

Furthermore, since TMs come with multiple advantages over state-of-the-art machine learning approaches, such as requiring a smaller memory footprint \cite{granmo2019convolutional, abeyrathna2020novel}, faster inference \cite{wheeldon2020tsetlin}, having fault-tolerant properties \cite{shafik2020explainability}, and interpretability \cite{wang2017bayesian, abeyrathna2021extending}, TM-based hardware has been implemented and rigorously tested, demonstrating their viability and efficacy in practical applications \cite{mao2022automated}. These hardware implementations capitalize on the inherent parallelism \cite{abeyrathna2021massively} and simplicity of TMs, making them well-suited for real-time and resource-constrained environments \cite{wheeldon2020learning}.

Despite the broad applicability of TMs across diverse domains and their practical advantages, there remains a gap in understanding how TMs quantify uncertainty in their predictions. Although some prior work has addressed uncertainty quantification with Tsetlin machines \cite{PredictionUncertainty, abeyrathna2023extension}, they are limited to continuous value prediction applications. In this study, we introduce the \textbf{\textit{Probabilistic Tsetlin Machine (PTM)}} as a novel framework for uncertainty quantification; a more robust, reliable, and interpretable approach, which can be adapted to many of the varieties of TMs mentioned above. We first discuss the basic TM theories and formally introduce the PTM in Section~\ref{sec2}. This covers the modifications needed on the TMs to obtain probabilistic outputs, and that is also the main contribution of this study. The importance of uncertainty quantification is discussed in Section~\ref{uq}. As explained in Section~\ref{experiments}, we conduct experiments using both artificial and real-world data. We conclude our findings in Section~\ref{conclusion}.

\section{Uncertainty Quantification}\label{uq}
\textbf{Overview:} Modeling uncertainty in scientific matters is not unique to artificial intelligence (AI) and is not just a contemporary concern. Figures like Huyghens, Pascal, Chevalier de Meré, and Jacques Bernoulli had mentioned the concept back in the 17th century in their work. Some challenges in knowledge representation and reasoning due to imprecision, uncertainty, and conflicting information were apparent with the rise of computer science. However, up until the 1980s, this was discussed separately from probability theory and decision-making concerns \cite{denoeux2020representations}.

In the past few decades, the potential of AI has become increasingly evident within the field of applied sciences \cite{abdar2022need}. However, AI systems need to estimate their uncertainty to make safe decisions in domains where mistakes are costly. Hence, for AI-based predictive models, uncertainty quantification serves as a measure of the model's confidence or reliability in making predictions under limited domain knowledge, incomplete or noisy data, or inherent randomness in the system and enhances the trust in their predictions \cite{seoni2023application}.

One of the many separations of uncertainty estimations involves separating uncertainty introduced by the AI model stemming from issues like a poor representation of the training data and insufficient knowledge, from uncertainty arising from the data itself due to reasons such as uninformative, incomplete, conflicting, or noisy data. These are called epistemic uncertainty and aleatoric uncertainty, respectively \cite{der2009aleatory}. Although epistemic uncertainty can be reduced by improving training and model architecture, aleatoric uncertainty arising from data is irreducible \cite{gawlikowski2021survey}.

Uncertainty can be quantified using two common methods: ensemble learning and probabilistic machine learning. The latter, also known as Bayesian machine learning, applies probability theory to extract information from data \cite{ghahramani2015}. However, probabilistic methods face computational challenges, especially when dealing with large or complex data or models, such as deep learning models. One can mitigate this issue by using approximation methods, such as variational inference and Markov Chain Monte Carlo, to estimate posterior distributions \cite{murphy2023}. However, one has to balance the trade-off between computation and accuracy when using these methods \cite{seoh2020}. Ensemble learning approaches, on the other hand, involve training several separate models and the final prediction is a collective effort of all. To estimate the uncertainty of the predictions, the models' divergence is utilized. Ensemble learning approaches also have associated limitations, namely, long training time, lack of interpretability, and sensitivity to noisy data and outliers.

To quantitatively assess the predictive accuracy and quantification of uncertainty of the models in scenarios with a discrete output space \( Y \), we employ several metrics: entropy, mutual information, and variance. These metrics are specifically chosen to measure the uncertainty of the models. For each test sample \( x_i \), these quantities are calculated as follows:

\textbf{Evaluation metrics:}

\textit{Predictive Mean}: For each test example, we generate $K$ prediction samples with each model. The \textit{Predictive Mean} represents the average predicted probabilities across the $K$ samples and serves as a point estimate of the target. It is calculated as follows:

\[
p(Y|X_i) = \frac{1}{K} \sum_{j=1}^{K} p_j(Y|X_i)
\]

Where $p_j(Y|X_i)$ is the $j$-th predicted probability for the $i$-th test example.

For any example \(X_i\), the method for computing \(p(Y|X)\) or \(p(X)\) differs across various models. In the PTM and Multilayer Perceptron with Monte Carlo dropout (MLP-MCd) models, \(p(X)\) is estimated by taking the average \(\bar{p}(X)\) of the sampled probabilities. In contrast, GPs can derive \(p(X)\) directly from the predictive distribution of the model. For Random Forests (RF), \(p(X)\) is typically computed as the mean of the probabilities predicted by each tree within the forest.

\textit{Predictive Entropy and mutual information}: They can be used to estimate the total and epistemic uncertainties respectively \cite{depeweg2018}. The total uncertainty can be quantified using the predictive entropy as follows:

\[
\hat{H}[p(Y|X_i)] = - \sum_{Y \in \mathcal{Y}} p(Y|X_i) \log_2 p(Y|X_i)
\]

The mutual information can be computed as follows: 

\[
\hat{I}[p(Y|X_i)] = \hat{H}[p(Y|X_i)] - \mathbb{E}_j[H(p_j(Y|X_i))]
\]

\textit{Standard Deviation}: It measures the spread or dispersion of the predictions and can be used as a measure of epistemic uncertainty. It is calculated as follows:

\[
\sigma_i = \sqrt{\frac{1}{K} \sum_{j=1}^{K} ({p}_{ij}(Y|X_i) - p(Y|X_i))^2}
\]

The \textit{Expected Calibration Error (ECE)} can be used as a measure of the reliability of a model's uncertainty estimates. It quantifies the difference between the model's predicted probabilities and the actual outcome. The ECE computes the weighted average of the absolute difference between accuracy and confidence across bins, measuring how well a model's predicted probabilities align with the true outcomes. It is calculated as follows: 
\[
\text{ECE} = \sum_{m=1}^{M} \frac{|B_m|}{D} \left| \text{acc}(B_m) - \text{conf}(B_m) \right|
\]
Where $M$ is the number of bins, $B_m$ is the set of samples in the $m^\text{th}$ bin, $D$ is the total number of samples, $acc(B_m)$ is the accuracy in bin $m$ calculated as the fraction of the correctly predicted samples, and $conf(B_m)$ is the mean predicted probability in bin $m$.


\section{Tsetlin Machines}\label{sec2}
\textbf{Structure:} Depending on the task at hand, a suitable Tsetlin Machine setup has to be used. For instance, if the goal is to classify samples into two classes, the basic TM designed for binary classification can be utilized. This TM is also the basis for most of the other versions of TMs, such as the multiclass Tsetlin machine (MTM) \cite{Ole1}, the regression Tsetlin machine (RTM) \cite{darshana2020regression}, and the convolutional Tsetlin machine (CTM) \cite{granmo2019convolutional}. In this paper, we use the basic TM for binary classification tasks (which we call TM) to explain the probabilistic Tsetlin machine (PTM), and the concept remains the same for all the other versions of the TMs.

The way TM addresses the pattern classification problem is, that it represents classes through a learned set of sub-patterns. In the TM, these sub-patterns are captured by clauses in the form of conjunctions of literals. TM receives features in binary form. Hence, a literal refers to either a propositional variable or its negation. TM uses a $m$ number of clauses to learn patterns, half of them recognizing patterns related to class $1$, while the rest learning patterns of class $0$.

Mathematically, a clause can be written as,
\begin{equation}\label{clauseout}
c_j = 1 \land \left(\bigwedge_{k \in {I}_j^I} x_k\right) \land \left( \bigwedge_{k \in \bar{{I}}_j^I} \lnot x_k\right).
\end{equation}

Here, $x_k$ and $\lnot x_k$ can be any input literal decided to be included in the clause $j$. The indexes of the nonnegated features to be included in the clause $j$ are recorded in ${I}_j^I$, while the indexes of the negated features that take part in the clause $j$ are stored in $\bar{I}_j^I$.

A dedicated team of Tsetlin automata (TAs) with $2N$ memory states \cite{narendra} decides the pattern a clause recognizes. More precisely, in each clause, each literal is assigned a TA. During the training, these TAs learn to \textit{include} or \textit{exclude} their literals in the clause. The $j^{th}$ clause, $c_j$ outputs 1 if all the included literals in the clause are 1s, otherwise, 0. 

\begin{table*}[!t]
\centering
\newcolumntype{P}[1]{>{\centering\arraybackslash}p{#1}}
\vspace{2mm}
\caption{Type I and Type II feedback to battle against false negatives and false positives.}\label{feedbacktable}
\begin{tabular}{P{8mm}|c|c|c|c|c|c|P{6mm}|P{6mm}|P{6mm}|P{6mm}}
\toprule
\multicolumn{3}{c|}{Feedback Type} & \multicolumn{4}{c|}{I} & \multicolumn{4}{c}{II}  \\ \hline
\multicolumn{3}{c|}{Clause Output} & \multicolumn{2}{c|}{1} & \multicolumn{2}{c|}{0} & \multicolumn{2}{c|}{1} & \multicolumn{2}{c}{0} \\ \hline
\multicolumn{3}{c|}{Literal Value} & 1 & 0 & 1 & 0 & 1 & 0 & 1 & 0 \\ \hline
\multirow{6}{*}{\rotatebox[origin=c]{90}{Current State}} & \multirow{3}{*}{Include} & Reward Probability & (s-1)/s & NA & 0 & 0 & 0 & NA & 0 & 0 \\
                        &                     & Inaction Probability & 1/s & NA & (s-1)/s & (s-1)/s & 1 & NA & 1 & 1\\
                        &                     & Penalty Probability & 0 & NA & 1/s & 1/s & 0 & NA & 0 & 0 \\ \cline{2-11}
                        & \multirow{3}{*}{Exclude} & Reward Probability & 0 & 1/s & 1/s & 1/s & 0 & 0 & 0 & 0\\
                        &                     & Inaction Probability & 1/s & (s-1)/s & (s-1)/s & (s-1)/s & 1 & 0 & 1 & 1\\
                        &                     & Penalty Probability & (s-1)/s & 0 & 0 & 0 & 0 & 1 & 0 & 0       \\ 
\bottomrule
\end{tabular}
\end{table*}

The number of clauses $m$ is a hyper-parameter set by the user. In the TM, clauses with odd indexes are given positive polarity, $c^+$ and they learn the patterns of class 1. The clauses with even indexes are given negative polarity, $c^-$ and they recognize the patterns related to class 0. The output of the TM is decided based on the difference between the patterns recognized by clauses from each class. For a specific set of features, if the clauses with positive polarity find more patterns than the clauses with negative polarity, the TM outputs 1, otherwise, 0.

\textbf{Learning:} The learning in TM involves guiding TAs in clauses to make the correct action. The TM is built around two types of feedback for this: Type I and Type II. Table~\ref{feedbacktable} summarizes how TA states should be updated based on Type I and Type II feedback. With both feedback types, the probabilities\footnote{The specificity $s$ ($s \geq 1$) is a hyper-parameter defined by the user and it controls the granularity of the sub-patterns.} for updating the TAs depend on the clause output (1 or 0), the literal value (1 or 0), and the current action of the TA (include or exclude). The Reward feedback strengthens the current action selected by the TA, the penalty feedback weakens the current action of the TA, and inaction keeps the state of the TA unchanged.

Type I feedback propels clauses with positive polarity outputs 1 when the training sample is from class 1, $y=1$ and clauses with negative polarity outputs 1 when the training sample is from class 0, $y=0$. Type I feedback does this by reinforcing the true positive outputs of clauses and erasing the patterns learned by clauses when they make false negative outputs. 

Type II feedback on the other hand combats the false positive clause outputs. Hence, Type II feedback is received by clauses with positive polarity when they output $1$ when the training sample is from class 0, $y=0$ and clauses with negative polarity when they output $1$ when the training sample is from class 1, $y=1$.  

\section{Probabilistic Tsetlin Machines}\label{sec2ptm}

In the PTM, our aim is to, instead of a single state update of TAs, consider the state of each TA to be represented by a distribution over all $2N$ potential states. We identify the state probability vector of the TA representing the $k^{th}$ literal in the $j^{th}$ clause as $SPV_{j,k} \in [0,1]^{2N}$. The learning process of PTM is summarized in Algorithm \ref{algo1}. As in the TM, the learning of the PTM starts by selecting the number of clauses $m$, the target $T$, and the specificity $s$. Unlike in the TM where the states of the TAs are randomly initialized to either $N^{th}$ or $N+1^{th}$ state, each state probability vector $SPV_{j,k}$ is initialized as, $p(S_N)=p(S_{N+1})=0.5 \quad \text{and} \quad p(S_i)=0 \quad \text{for} \quad i\neq N, N+1$, with $p(S_i)$ being the probability of staying at state $i$. 

In each learning step $n$, literals included in the $j^{th}$ clause are decided by sampling the states for TAs from their corresponding $SPV_{j,k}$ (also explained in Algorithm~\ref{clause}). The candidate clauses to receive Type~I feedback are selected stochastically with probability $\frac{T - \mathrm{max}(-T, \mathrm{min}(T, v))}{2T}$ and those to receive Type~II feedback are selected stochastically with probability $\frac{T + \mathrm{max}(-T, \mathrm{min}(T, v))}{2T}$. Here, the target $T$, specified by the user, determines the number of clauses involved in learning each sub-pattern.

We still utilize the Type I and Type II feedback in the training phase of PTM. However, since we need to update the probability vector $SPV_{j,k}$, instead of making state transitions depending on the type of feedback TAs receive, we convert the feedback probability tables into equivalent Transition Probability Matrices (TPMs). In other words, we create four TPMs that represent the unique columns in Table~\ref{feedbacktable}. These PTMs can be seen from Table~\ref{TPM1} to Table~\ref{TPM4} where $TPM_1$ equivalent to Column 1 of the Type I feedback table, $TPM_2$ equivalent to Column 2 of the Type I feedback table, $TPM_3$ equivalent to Column 3 and 4 of the Type I feedback table, and $TPM_4$ equivalent to Column 2 of the Type II feedback table. However, we do not need a TPM for the remaining columns in Type II feedback tables since, with those columns, the $SPV_{j,k}$ will not be altered.

\begin{algorithm*}[!t]
\caption{Learning process}\label{algo1}
\begin{algorithmic}[1]
\STATE \textbf{Initialize TM:} $m$, $T$, $s$, Set the initial state probabilities of TAs: $p(S_N) = p(S_{N+1}) = 0.5$
\STATE \textbf{Convert feedback tables into transition probability matrix}
\STATE \textbf{Begin:} $n^{th}$ training round
\STATE \textbf{Compute:} Clause outputs (\textbf{Algorithm \ref{clause}})
\FOR{$j = 1, ...,m$} 
    \IF{satisfy condition with probability $\frac{T + \mathrm{max}(-T, \mathrm{min}(T, v))}{2T}$ or $\frac{T - \mathrm{max}(-T, \mathrm{min}(T, v))}{2T}$}
        \IF{($y = 1$ \textbf{and} $j$ is odd) \textbf{or} ($y = 0$ \textbf{and} $j$ is even)}
            \STATE Update $SPV$ with \textbf{Algorithm \ref{typeI}}
        \ELSIF{ ($y = 1$ \textbf{and} $j$ is even) \textbf{or} ($y = 0$ \textbf{and} $j$ is odd)} 
            \STATE Update $SPV$ with \textbf{Algorithm \ref{typeII}}
        \ENDIF
    \ENDIF
\ENDFOR
\end{algorithmic}
\end{algorithm*}

Algorithms \ref{typeI} and \ref{typeII} describe how Type I and Type II feedbacks update the $SPV_{j,k}$ using the corresponding TPMs, respectively. For instance, when it is Type I feedback, clause output is 1, and the literal value is 1, the $SPV_{j,k}$ is updated by multiplying the vector by ${TPM}_1$. This increases the probability of the "include" action of the TA instead of shifting the state towards the 'include' action by one step. This simplifies the implementation of TMs as it skips some of the \textbf{$IF-THEN$} conditions. This way, at the end of the training, we end up having distributions over the states of all the TAs.

\textbf{Inference:} After the training phase concludes, the PTM is ready to generate predictions on previously unseen samples. During the inference phase, each time an inquiry is directed towards the PTM, the TA states for all the TAs in all the clauses are sampled based on the state probability vectors corresponding to each TA. These states determine the actions of their TAs and hence also the clause outputs. Consequently, both the outputs of the clauses and that of the PTM itself may exhibit variation, even when the identical input sample is repeatedly presented to the PTM, similar to how Bayesian Neural Networks make predictions with the learned distributions of its parameters.

\begin{table*}[!t]
\centering
\caption{Column 1 of Type I feedback table: Transition probability matrix-1 ($TPM_1$)} \label{TPM1}
\begin{tabular}{cccccccc}
                             &                                               &                                 & \multicolumn{5}{c}{Time at $t-1$}                                                                                                                                           \\
                             &                                               &                                 & \multicolumn{2}{c}{Ex}                                               &                          & \multicolumn{2}{c}{In}                                          \\ \cline{3-8} 
                             & \multicolumn{1}{c|}{}                         & \multicolumn{1}{c|}{}           & \multicolumn{1}{c|}{$S_1$}           & \multicolumn{1}{c|}{$S_2$}         & \multicolumn{1}{c|}{...} & \multicolumn{1}{c|}{$S_{2N-1}$}      & \multicolumn{1}{c|}{$S_{2N}$} \\ \cline{3-8} 
\multirow{5}{*}{\rotatebox[origin=c]{90}{Time at $t$}} & \multicolumn{1}{c|}{\multirow{2}{*}{Ex}} & \multicolumn{1}{c|}{$S_1$}      & \multicolumn{1}{c|}{$\frac{1}{s}$}   & \multicolumn{1}{c|}{0}             & \multicolumn{1}{c|}{...} & \multicolumn{1}{c|}{0}               & \multicolumn{1}{c|}{0}        \\ \cline{3-8} 
                             & \multicolumn{1}{c|}{}                         & \multicolumn{1}{c|}{$S_2$}      & \multicolumn{1}{c|}{$\frac{s-1}{s}$} & \multicolumn{1}{c|}{$\frac{1}{s}$} & \multicolumn{1}{c|}{...} & \multicolumn{1}{c|}{0}               & \multicolumn{1}{c|}{0}        \\ \cline{3-8} 
                             & \multicolumn{1}{c|}{}                         & \multicolumn{1}{c|}{...}        & \multicolumn{1}{c|}{...}             & \multicolumn{1}{c|}{...}           & \multicolumn{1}{c|}{...} & \multicolumn{1}{c|}{...}             & \multicolumn{1}{c|}{...}      \\ \cline{3-8} 
                             & \multicolumn{1}{c|}{\multirow{2}{*}{In}} & \multicolumn{1}{c|}{$S_{2N-1}$} & \multicolumn{1}{c|}{0}               & \multicolumn{1}{c|}{0}             & \multicolumn{1}{c|}{...} & \multicolumn{1}{c|}{$\frac{1}{s}$}   & \multicolumn{1}{c|}{0}        \\ \cline{3-8} 
                             & \multicolumn{1}{c|}{}                         & \multicolumn{1}{c|}{$S_{2N}$}   & \multicolumn{1}{c|}{0}               & \multicolumn{1}{c|}{0}             & \multicolumn{1}{c|}{...} & \multicolumn{1}{c|}{$\frac{s-1}{s}$} & \multicolumn{1}{c|}{1}        \\ \cline{3-8} 
\end{tabular}
\end{table*}

\begin{table*}[!t]
\centering
\caption{Column 2 of Type I feedback table: Transition probability matrix-2 ($TPM_2$)}\label{TPM2}
\begin{tabular}{cccccccc}
                             &                                               &                                 & \multicolumn{5}{c}{Time at $t-1$}                                                                                                                              \\
                             &                                               &                                 & \multicolumn{2}{c}{Ex}                                       &                          & \multicolumn{2}{c}{In}                                     \\ \cline{3-8} 
                             & \multicolumn{1}{c|}{}                         & \multicolumn{1}{c|}{}           & \multicolumn{1}{c|}{$S_1$} & \multicolumn{1}{c|}{$S_2$}           & \multicolumn{1}{c|}{...} & \multicolumn{1}{c|}{$S_{2N-1}$} & \multicolumn{1}{c|}{$S_{2N}$} \\ \cline{3-8} 
\multirow{5}{*}{\rotatebox[origin=c]{90}{Time at $t$}} & \multicolumn{1}{c|}{\multirow{2}{*}{Ex}} & \multicolumn{1}{c|}{$S_1$}      & \multicolumn{1}{c|}{1}     & \multicolumn{1}{c|}{$\frac{1}{s}$}   & \multicolumn{1}{c|}{...} & \multicolumn{1}{c|}{0}          & \multicolumn{1}{c|}{0}        \\ \cline{3-8} 
                             & \multicolumn{1}{c|}{}                         & \multicolumn{1}{c|}{$S_2$}      & \multicolumn{1}{c|}{0}     & \multicolumn{1}{c|}{$\frac{s-1}{s}$} & \multicolumn{1}{c|}{...} & \multicolumn{1}{c|}{0}          & \multicolumn{1}{c|}{0}        \\ \cline{3-8} 
                             & \multicolumn{1}{c|}{}                         & \multicolumn{1}{c|}{...}        & \multicolumn{1}{c|}{...}   & \multicolumn{1}{c|}{...}             & \multicolumn{1}{c|}{...} & \multicolumn{1}{c|}{...}        & \multicolumn{1}{c|}{...}      \\ \cline{3-8} 
                             & \multicolumn{1}{c|}{\multirow{2}{*}{In}} & \multicolumn{1}{c|}{$S_{2N-1}$} & \multicolumn{1}{c|}{0}     & \multicolumn{1}{c|}{0}               & \multicolumn{1}{c|}{...} & \multicolumn{1}{c|}{1}          & \multicolumn{1}{c|}{0}        \\ \cline{3-8} 
                             & \multicolumn{1}{c|}{}                         & \multicolumn{1}{c|}{$S_{2N}$}   & \multicolumn{1}{c|}{0}     & \multicolumn{1}{c|}{0}               & \multicolumn{1}{c|}{...} & \multicolumn{1}{c|}{0}          & \multicolumn{1}{c|}{1}        \\ \cline{3-8} 
\end{tabular}
\end{table*}

\begin{table*}[!t]
\centering
\caption{Column 3 and 4 of Type I feedback table: Transition probability matrix-3 ($TPM_3$)}\label{TPM3}
\begin{tabular}{cccccccc}
                             &                                               &                                 & \multicolumn{5}{c}{Time at $t-1$}                                                                                                                                          \\
                             &                                               &                                 & \multicolumn{2}{c}{Ex}                                       &                          & \multicolumn{2}{c}{In}                                                 \\ \cline{3-8} 
                             & \multicolumn{1}{c|}{}                         & \multicolumn{1}{c|}{}           & \multicolumn{1}{c|}{$S_1$} & \multicolumn{1}{c|}{$S_2$}           & \multicolumn{1}{c|}{...} & \multicolumn{1}{c|}{$S_{2N-1}$}      & \multicolumn{1}{c|}{$S_{2N}$}        \\ \cline{3-8} 
\multirow{5}{*}{\rotatebox[origin=c]{90}{Time at $t$}} & \multicolumn{1}{c|}{\multirow{2}{*}{Ex}} & \multicolumn{1}{c|}{$S_1$}      & \multicolumn{1}{c|}{1}     & \multicolumn{1}{c|}{$\frac{1}{s}$}   & \multicolumn{1}{c|}{...} & \multicolumn{1}{c|}{0}               & \multicolumn{1}{c|}{0}               \\ \cline{3-8} 
                             & \multicolumn{1}{c|}{}                         & \multicolumn{1}{c|}{$S_2$}      & \multicolumn{1}{c|}{0}     & \multicolumn{1}{c|}{$\frac{s-1}{s}$} & \multicolumn{1}{c|}{...} & \multicolumn{1}{c|}{0}               & \multicolumn{1}{c|}{0}               \\ \cline{3-8} 
                             & \multicolumn{1}{c|}{}                         & \multicolumn{1}{c|}{...}        & \multicolumn{1}{c|}{...}   & \multicolumn{1}{c|}{...}             & \multicolumn{1}{c|}{...} & \multicolumn{1}{c|}{...}             & \multicolumn{1}{c|}{...}             \\ \cline{3-8} 
                             & \multicolumn{1}{c|}{\multirow{2}{*}{In}} & \multicolumn{1}{c|}{$S_{2N-1}$} & \multicolumn{1}{c|}{0}     & \multicolumn{1}{c|}{0}               & \multicolumn{1}{c|}{...} & \multicolumn{1}{c|}{$\frac{s-1}{s}$} & \multicolumn{1}{c|}{$\frac{1}{s}$}   \\ \cline{3-8} 
                             & \multicolumn{1}{c|}{}                         & \multicolumn{1}{c|}{$S_{2N}$}   & \multicolumn{1}{c|}{0}     & \multicolumn{1}{c|}{0}               & \multicolumn{1}{c|}{...} & \multicolumn{1}{c|}{0}               & \multicolumn{1}{c|}{$\frac{s-1}{s}$} \\ \cline{3-8} 
\end{tabular}
\end{table*}

\begin{table*}[!t]
\centering
\caption{Column 2 of Type II feedback table: Transition probability matrix-4 ($TPM_4$)}\label{TPM4}
\begin{tabular}{cccccccc}
                             &                                               &                                 & \multicolumn{5}{c}{Time at $t-1$}                                                                                                                    \\
                             &                                               &                                 & \multicolumn{2}{c}{Ex}                             &                          & \multicolumn{2}{c}{In}                                     \\ \cline{3-8} 
                             & \multicolumn{1}{c|}{}                         & \multicolumn{1}{c|}{}           & \multicolumn{1}{c|}{$S_1$} & \multicolumn{1}{c|}{$S_2$} & \multicolumn{1}{c|}{...} & \multicolumn{1}{c|}{$S_{2N-1}$} & \multicolumn{1}{c|}{$S_{2N}$} \\ \cline{3-8} 
\multirow{5}{*}{\rotatebox[origin=c]{90}{Time at $t$}} & \multicolumn{1}{c|}{\multirow{2}{*}{Ex}} & \multicolumn{1}{c|}{$S_1$}      & \multicolumn{1}{c|}{0}     & \multicolumn{1}{c|}{0}     & \multicolumn{1}{c|}{...} & \multicolumn{1}{c|}{0}          & \multicolumn{1}{c|}{0}        \\ \cline{3-8} 
                             & \multicolumn{1}{c|}{}                         & \multicolumn{1}{c|}{$S_2$}      & \multicolumn{1}{c|}{1}     & \multicolumn{1}{c|}{0}     & \multicolumn{1}{c|}{...} & \multicolumn{1}{c|}{0}          & \multicolumn{1}{c|}{0}        \\ \cline{3-8} 
                             & \multicolumn{1}{c|}{}                         & \multicolumn{1}{c|}{...}        & \multicolumn{1}{c|}{...}   & \multicolumn{1}{c|}{...}   & \multicolumn{1}{c|}{...} & \multicolumn{1}{c|}{...}        & \multicolumn{1}{c|}{...}      \\ \cline{3-8} 
                             & \multicolumn{1}{c|}{\multirow{2}{*}{In}} & \multicolumn{1}{c|}{$S_{2N-1}$} & \multicolumn{1}{c|}{0}     & \multicolumn{1}{c|}{0}     & \multicolumn{1}{c|}{...} & \multicolumn{1}{c|}{1}          & \multicolumn{1}{c|}{0}        \\ \cline{3-8} 
                             & \multicolumn{1}{c|}{}                         & \multicolumn{1}{c|}{$S_{2N}$}   & \multicolumn{1}{c|}{0}     & \multicolumn{1}{c|}{0}     & \multicolumn{1}{c|}{...} & \multicolumn{1}{c|}{0}          & \multicolumn{1}{c|}{1}        \\ \cline{3-8} 
\end{tabular}
\end{table*}

\begin{algorithm}[!t]
\caption{Compute clause output}\label{clause}
\begin{algorithmic}[1]
\FOR{clause $j=1,...,m$}
    \STATE ${I}_j^I \leftarrow$ \{ \}
    \STATE $\bar{I}_j^I \leftarrow$ \{ \}
    \FOR{feature $k=1,...,2o$}
        \STATE Sample a state, $S_{j,k}$ based on $SPV_{j,k}$
        \IF{$k \le o$ \textbf{and} $S_{j,k} > S_N$}
        \STATE ${I}_j^I$.insert(k)
        \ELSIF { $k > o$ \textbf{and} $S_{j,k} > S_N$}
        \STATE $\bar{I}_j^I$.insert(k)
        \ENDIF
    \ENDFOR
    \STATE Compute output of clause $j$
\ENDFOR
\end{algorithmic}
\end{algorithm}

\begin{algorithm}[!t]
\caption{State probability update due to Type I feedback}\label{typeI}
\begin{algorithmic}[1]
\IF{$c_j = 1$} 
    \FOR{feature $k=1,...,2o$}
        \IF{$x'_{k} = 1$}
            \STATE $SPV_{j,k}(n+1) \leftarrow SPV_{j,k}(n) \cdot {TPM}_1$
        \ELSIF{ $x'_{k} = 0$} 
            \STATE $SPV_{j,k}(n+1) \leftarrow SPV_{j,k}(n) \cdot {TPM}_2$
        \ENDIF
    \ENDFOR
\ELSIF{ $c_j = 0$}
    \STATE $SPV_{j,k}(n+1) \leftarrow SPV_{j,k}(n) \cdot {TPM}_3$
\ENDIF
\end{algorithmic}
\end{algorithm}

\begin{algorithm}[!t]
\caption{State probability update due to Type II feedback}\label{typeII}
\begin{algorithmic}[1]
\IF{$c_j = 1$} 
    \FOR{feature $k=1,...,2o$}
        \IF{$x'_{k} = 0$}
            \STATE $SPV_{j,k}(n+1) \leftarrow SPV_{j,k}(n) \cdot {TPM}_4$
        \ELSIF{ $x'_{k} = 1$} 
            \STATE $SPV_{j,k}(n+1) \leftarrow SPV_{j,k}(n)$
        \ENDIF
    \ENDFOR
\ELSIF{ $c_j = 0$} 
    \STATE $SPV_{j,k}(n+1) \leftarrow SPV_{j,k}(n)$
\ENDIF
\end{algorithmic}
\end{algorithm}


\section{Experiments and Results}\label{experiments}

This section describes the experiments we conducted and the results we obtained. First, we introduce the benchmark methods used for comparison with the proposed PTM. Then, we illustrate what the probability vectors for different TAs look like for the noisy-XOR dataset. Finally, we use both synthetic and real-world data to demonstrate how accurately PTM can quantify uncertainties compared to the benchmark methods.

\subsection{Benchmark Methods}

We compare our proposed PTM with different machine learning techniques for uncertainty quantification:
\begin{itemize}
    \item \textbf{Gaussian Process (GP)}: A GP \cite{williams2006} is a Bayesian, non-parametric learning method for regression and classification problems. It models the probability distribution over possible functions based on observed data. The GP is defined by a mean function and a covariance or kernel function, which determine the similarity between data points. 
    \item \textbf{Multilayer Perceptron with Monte Carlo dropout (MLP-MCd)}: MLP-MCd is a neural network technique that integrates dropout layers not only during training for regularization but also during inference for uncertainty estimation \cite{gal2016}. By performing multiple forward passes with random neuron deactivation, the MLP generates a distribution of outputs offering insights into the model's uncertainty.
    \item \textbf{Random Forests (RF)}: RF \cite{breiman2001} is an ensemble machine learning technique that involves constructing multiple decision trees and combining their predictions to obtain an output with a measure of confidence. RF operates by using a technique called bagging, which helps to reduce variance without increasing bias. Furthermore, RF utilizes random subsets of features at each split point in the tree-building process.
\end{itemize}

\subsection{Steady state probabilities (SSPs) for the noisy-XOR dataset}
To illustrate what the probability vectors look like, we utilize a variant of the noisy-XOR dataset, initially introduced in \cite{Ole1}. In this adapted version, we eliminated redundant features and introduced a 30\% random inversion in the XOR outputs. Subsequently, we task the PTM with learning the distinctive patterns for classes 1 and 0 using merely 4 clauses, $m=4$, with each clause dedicated to a specific pattern.

\begin{figure*}[!ht]
     \centering
     \begin{subfigure}[b]{0.99\textwidth}
         \includegraphics[width=\textwidth]{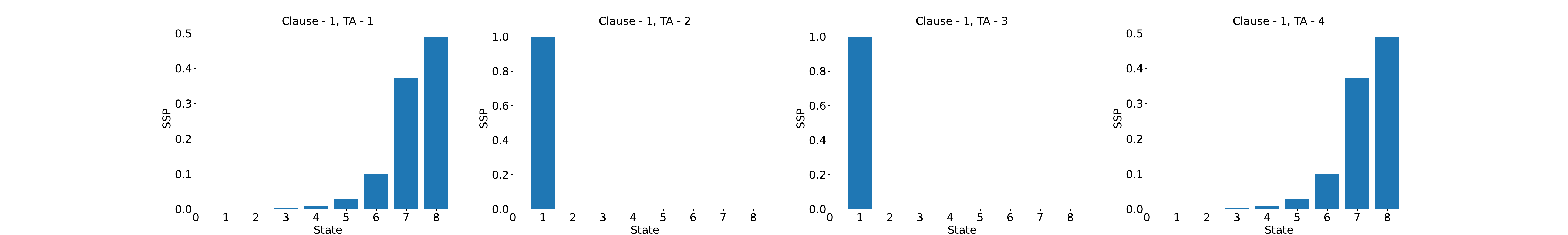}
     \end{subfigure}
     \hfill
     \begin{subfigure}[b]{0.99\textwidth}
         \includegraphics[width=\textwidth]{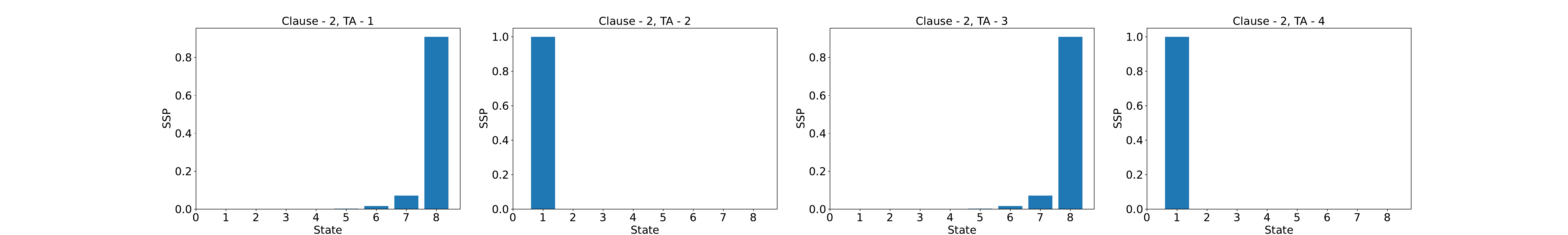}
     \end{subfigure}
     \hfill
     \begin{subfigure}[b]{0.99\textwidth}
         \includegraphics[width=\textwidth]{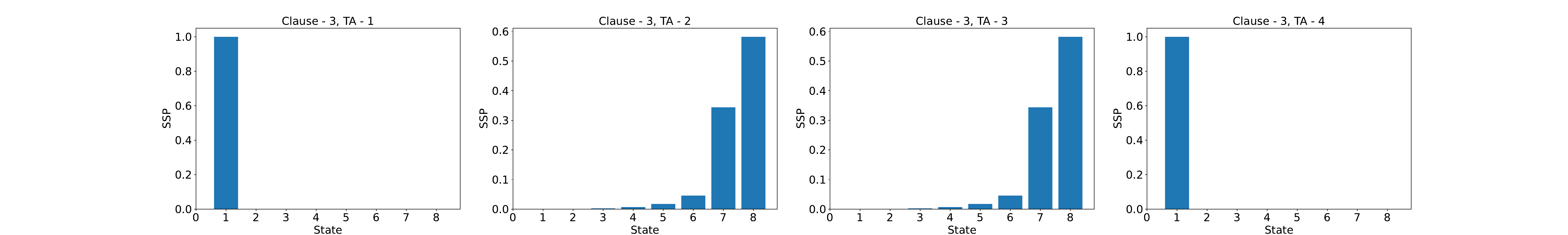}
     \end{subfigure}
    \hfill
     \begin{subfigure}[b]{0.99\textwidth}
         \includegraphics[width=\textwidth]{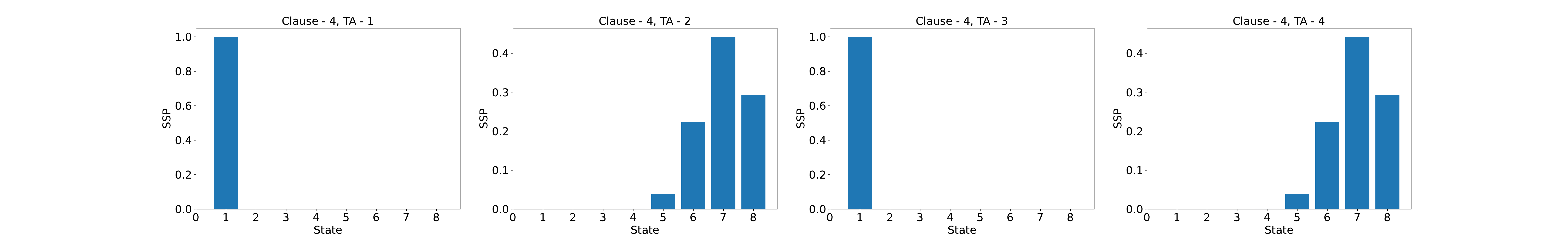}
     \end{subfigure}
\caption{SSPs of TAs in clauses to tackle the noise-free XOR problem}\label{XOR}
\end{figure*}

\begin{figure*}[!ht]
     \centering
     \begin{subfigure}[b]{0.99\textwidth}
         \includegraphics[width=\textwidth]{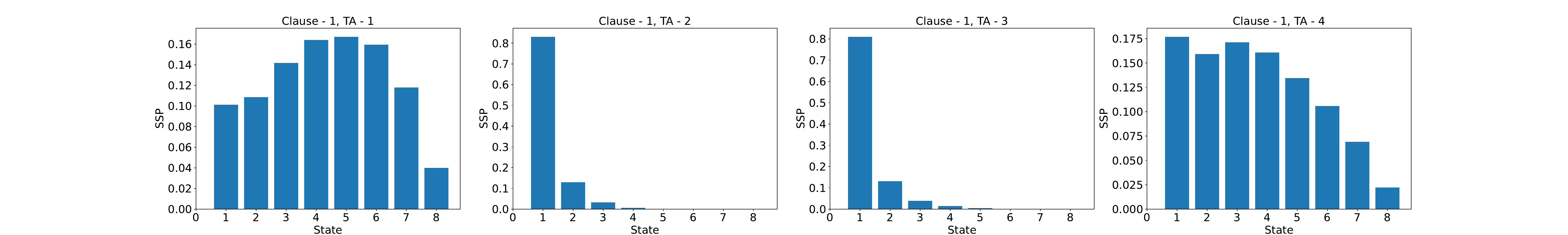}
     \end{subfigure}
     \hfill
     \begin{subfigure}[b]{0.99\textwidth}
         \includegraphics[width=\textwidth]{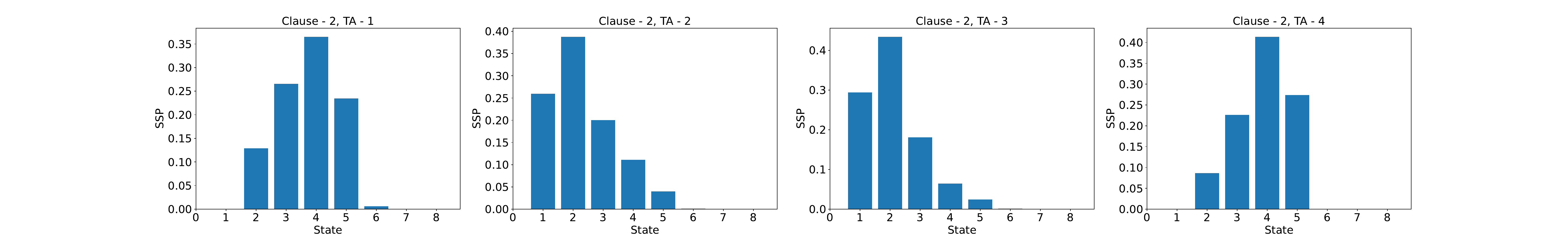}
     \end{subfigure}
     \hfill
     \begin{subfigure}[b]{0.99\textwidth}
         \includegraphics[width=\textwidth]{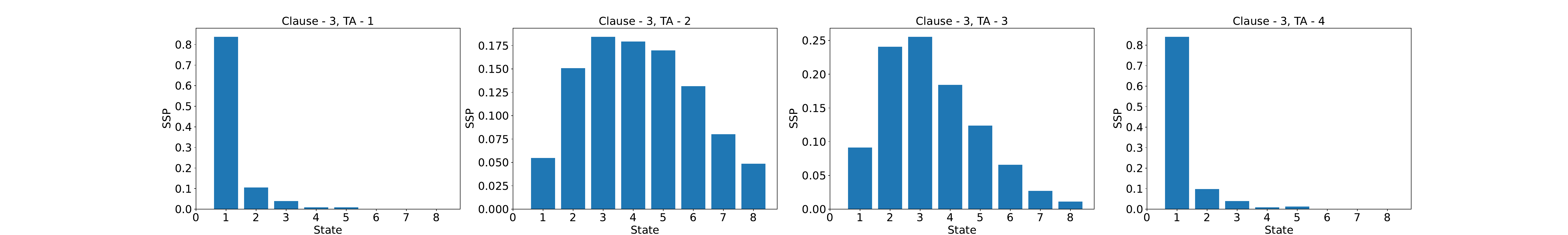}
     \end{subfigure}
    \hfill
     \begin{subfigure}[b]{0.99\textwidth}
         \includegraphics[width=\textwidth]{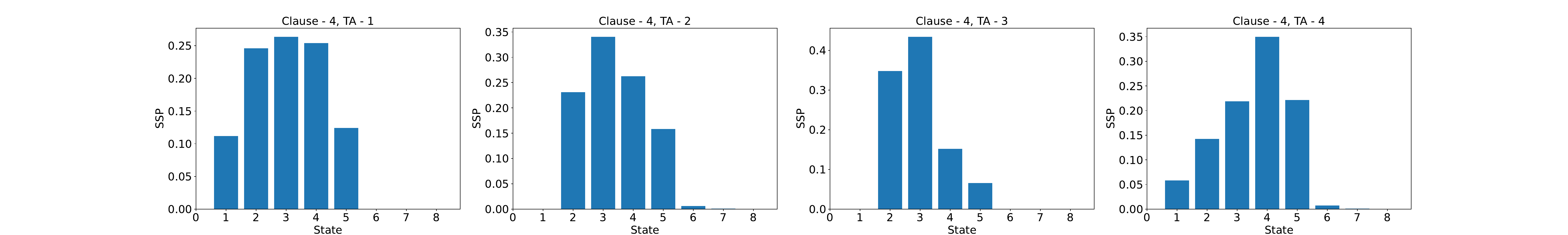}
     \end{subfigure}
     \vspace{1mm}
\caption{SSPs of TAs in clauses to tackle the noisy-XOR problem}\label{fig:three graphs} \label{noXOR}
\end{figure*}

\begin{figure*}[!t]
    \centering
    \includegraphics[width=0.65\linewidth]{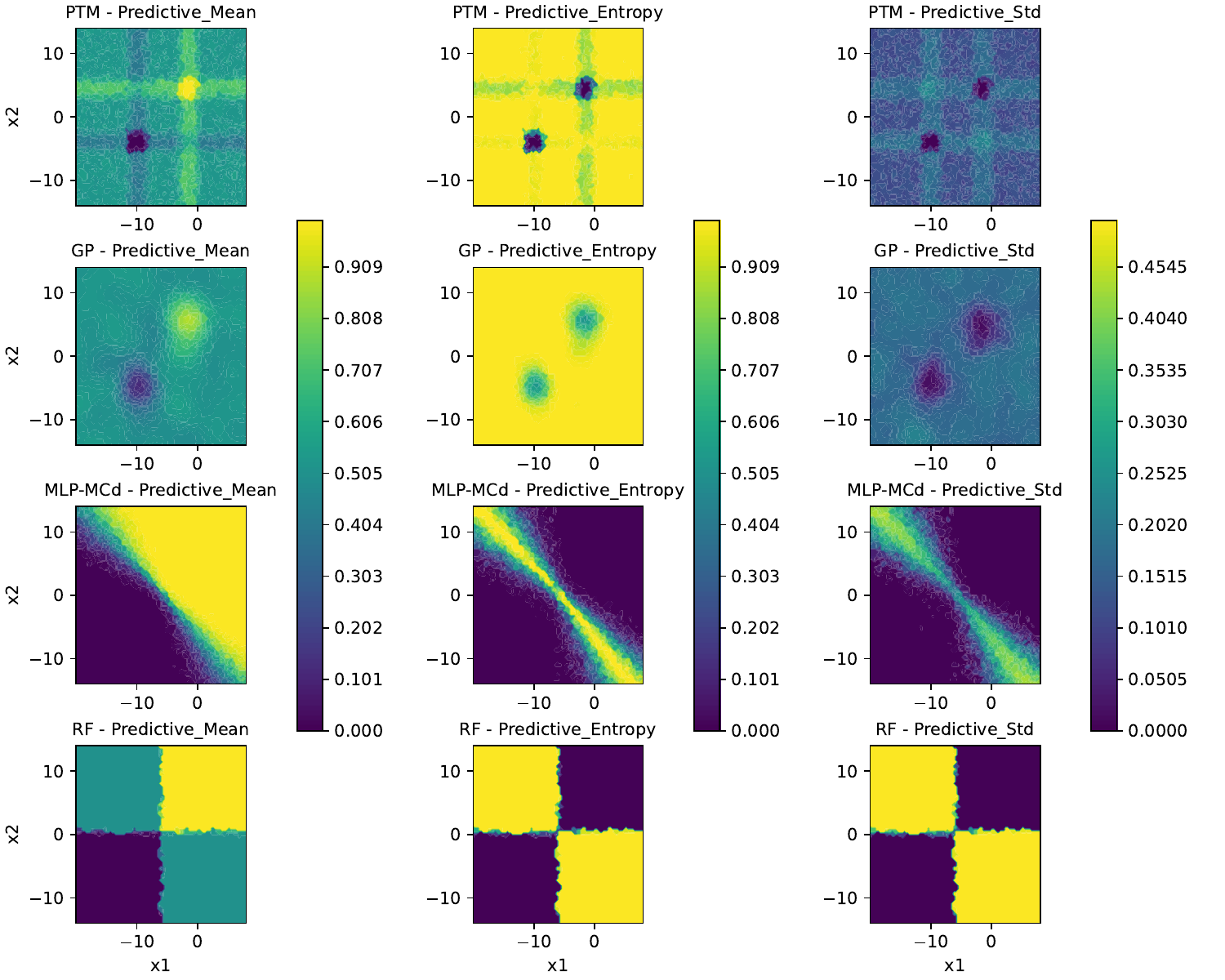}
    \caption{Predictive mean, entropy, and standard deviation of the PTM, GP, MLP-MCd, and RF models on the test data.}\label{fig:binary_results}
\end{figure*}

\begin{figure}[!t]
  \centering
  \includegraphics[width=0.5\linewidth]{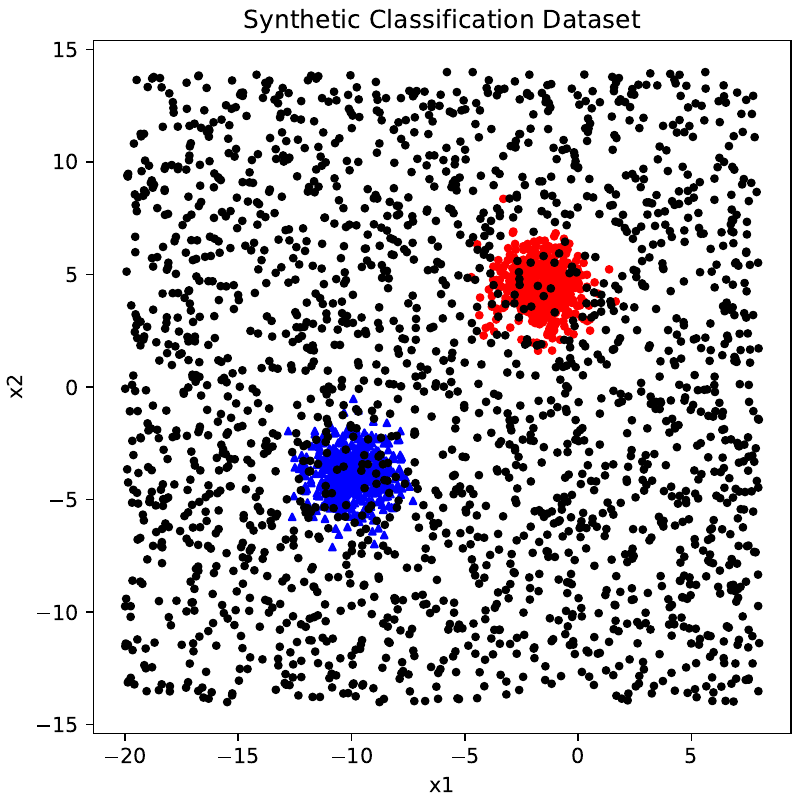}
  \caption{Synthetic binary classification dataset.}\label{fig:binary_data}
\end{figure}

For reference, the distributions in the absence of noise are depicted in Figure~\ref{XOR}. Here, TA-1, TA-2, TA-3, and TA-4 make decisions regarding inclusion and exclusion of the literals $x_1, \lnot x_1, x_2$, and $\lnot x_2$, respectively. As anticipated, $c_1$ and $c_3$ discern the patterns for class 1, while $c_2$ and $c_4$ learn those for class 0. For instance, $c_1$ grasps the pattern (0, 1) by including $\lnot x_1$ and $x_2$ in the clause, thereby yielding an output of 1 solely when $x_1$ is 0 and $x_2$ is 1 in the sample.

In contrast to the clear and straightforward inclusions and exclusions of literals in clauses observed when the data is noise-free, the TAs adapt their representations to accommodate the uncertainty inherent in noisy data, as evident in Figure~\ref{noXOR}. While one might assume, based on prior knowledge about clause allocation to the classes, that $c_1$ has learned the pattern (0, 1) and $c_3$ has learned the pattern (1, 0), the precise composition of the patterns can vary. In other words, even though it's apparent that $c_1$ includes $\lnot x_1$ and $x_2$, and $c_3$ includes $x_1$ and $\lnot x_2$ in their clauses with higher probabilities, the pattern can be disturbed depending upon the decisions of the other TAs due to the shared probabilities among TA states of them.

\subsection{Illustration using a synthetic dataset}

Our experimental analysis assesses the models using a simple simulated dataset comprising 1000 labeled examples. Each data point in this dataset consists of two features, denoted as $X_1$ and $X_2$, and is assigned one of two labels: $0$ or $1$. To evaluate the models and gain insights into their performance, we generate 2000 additional unlabeled test examples. These test examples are randomly distributed over a larger range than the training data points. The dataset is represented in Figure \ref{fig:binary_data}. 

In Figure \ref{fig:binary_results}, a comparison of the PTM's predictions and uncertainty quantification with other benchmark models is presented. As explained earlier, the predictive mean, entropy, and standard deviation are calculated for each test point using 100 prediction samples. Each panel represents the interpolated surfaces of the values obtained. The predictive mean values indicate that all models effectively differentiate between the two classes. Notably, the PTM and GP models accurately capture the distribution of both classes. The PTM, in particular, delimits some decision boundaries relative to the input values. The MLP-MCd model implements a linear classifier, whereas the RF model partitions the input space into four distinct regions.

The total uncertainty, as indicated by entropy, from both the PTM and GP, is high outside the regions covered by the training data. In contrast, the MLP-MCd model exhibits high uncertainty only along the decision boundary and very low uncertainty otherwise. The RF model divides the input space into four areas. It shows high uncertainty in regions with no training data and low uncertainty in areas with training data. However, unlike the PTM and GP, the RF is confident in regions that extend beyond the actual coverage of the training data.

\begin{figure}[!htb]
    \centering
    \includegraphics[width=0.5\linewidth]{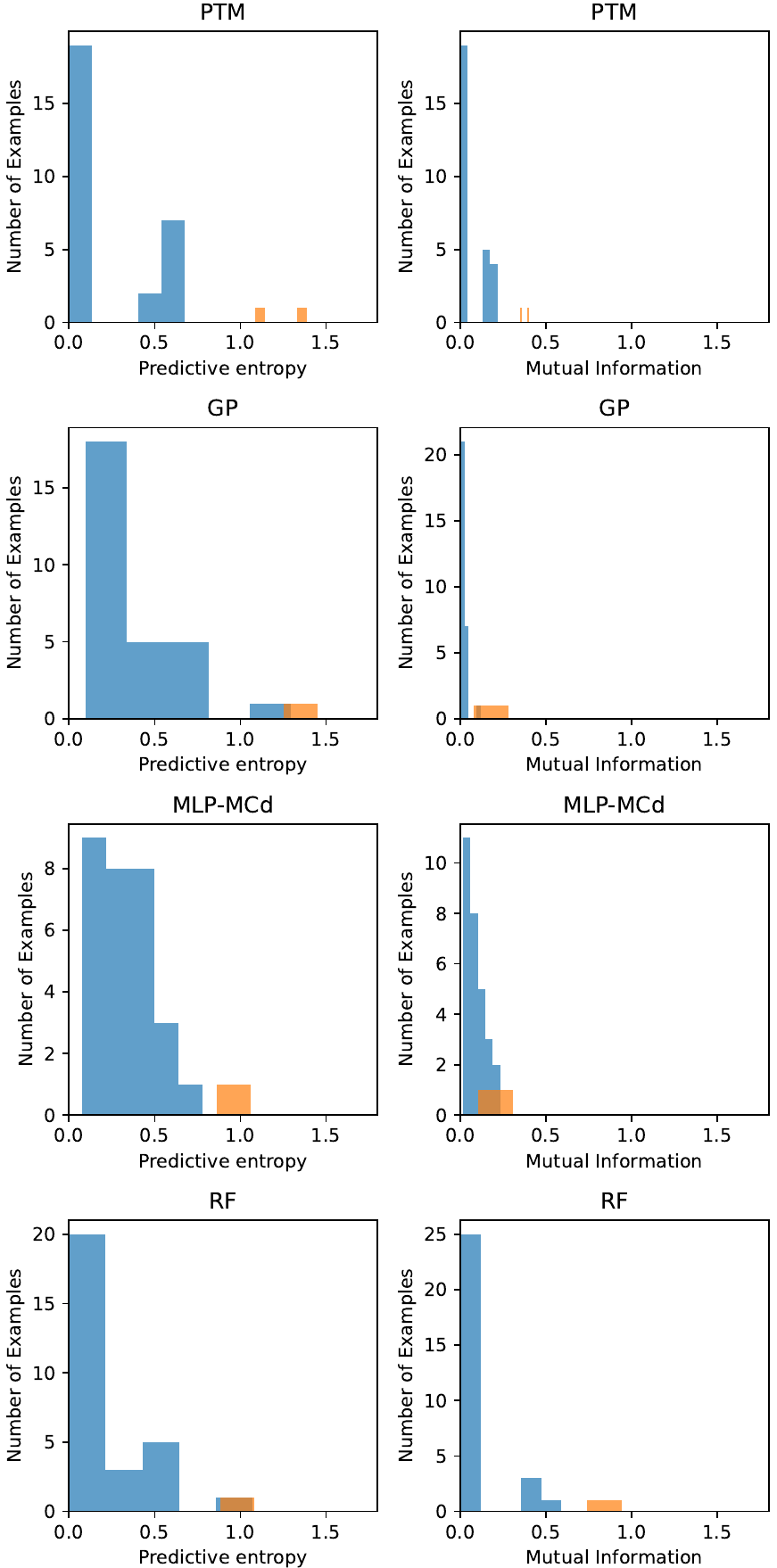}
    \caption{Predictive entropy and mutual information for correct and incorrect classifications of iris test dataset for the PTM, GP, MLP-MCd, and RF models.}\label{fig:multiclass_results}
\end{figure}

Regarding epistemic uncertainty, as indicated by the standard deviation, the PTM and GP demonstrate low uncertainty within regions covered by training data. Also, the models provide high uncertainty in sparsely populated areas, aligning with expectations that uncertainty should increase where data is scarce. Notably, the PTM reveals some patterns of uncertainty in no-data regions, with the highest uncertainty observed in ambiguous zones where feature values are associated with conflicting classes. Uncertainty diminishes in regions predominantly influenced by a single class. The MLP-MCd model exhibits high uncertainty along its linear decision boundaries yet displays confidence in areas uncovered by training data. The RF model exhibits low uncertainty in blocks with training data and high uncertainty in the other regions. However, it can be very confident in areas not covered by training data.

We conclude that the uncertainty quantification offered by the MLP-MCd model can be misleading, as it shows high confidence in no-data areas. In addition, although the RF model effectively discriminates between the two classes, it occasionally presents low uncertainty in areas where higher uncertainty would be expected. The GP and PTM models provide good uncertainty quantification, with the PTM providing a particularly interpretable uncertainty. This analysis emphasizes the importance of choosing appropriate models for accurate uncertainty estimation in predictive tasks. 

\subsection{Multiclass classification using the Iris dataset}

In this section, we evaluate the performance of the multiclass PTM on the Iris dataset, comparing it with the GP, MLP-MCd, and RF. We allocate 80\% of the dataset for training the models and the remaining 20\% for testing. The evaluation metrics include predictive entropy, mutual information, and ECE.

Figure \ref{fig:multiclass_results} illustrates the results, highlighting that both predictive entropy and mutual information are significantly higher for incorrect predictions, which aligns with the expected behavior of a good probabilistic model. Table \ref{tab:ece} presents the ECE for each model, where lower values indicate better model calibration. The ECE values in the table indicate effective calibration across all models. 

\begin{table}[h]
\caption{ECE on the Iris test dataset.}\label{tab:ece}
\centering
\begin{tabular}{l|l} 
\toprule
Model & ECE \\
\toprule
PTM & 0.0099 \\ \hline
GP & 0.0063 \\ \hline
MLP-MCd & 0.0100\\ \hline
RF & 0.0074\\
\bottomrule
\end{tabular}
\end{table}

\vspace{-5mm}
\section{Conclusion} \label{conclusion}
In conclusion, the Probabilistic Tsetlin Machine (PTM) framework introduced in this paper offers a promising solution to address the challenge of uncertainty quantification in predictive tasks with Tsetlin Machines. Through experimental analysis, we have demonstrated the effectiveness of the PTM in accurately quantifying uncertainties. The experiments conducted on both simulated and real-world datasets showcase the PTM's competitive performance in terms of predictive entropy and expected calibration error. Moving forward, further exploration of the PTM's capabilities and its application across diverse datasets and domains could yield valuable insights into its potential for addressing complex predictive challenges while providing transparent uncertainty estimates. When utilized to its full potential, the PTM not only computes uncertainty estimations but could also provide reasons for higher and lower uncertainties, making it a valuable asset in various domains where wrong predictions are costly.
 





\end{document}